\documentclass[a4paper]{article}

\usepackage{textcomp}
\usepackage[utf8]{inputenc}
\usepackage[T1]{fontenc}
\usepackage{newunicodechar}
\newunicodechar{’}{\textquoteright}
\newunicodechar{–}{\textendash}
\newunicodechar{ř}{\v{r}}
\newunicodechar{š}{\v{s}}
\usepackage[round]{natbib}
\usepackage[hyphens]{url}
\usepackage{hyperref}
\usepackage[dvipsnames]{xcolor}
\usepackage{amsfonts}
\usepackage{amssymb}
\usepackage{amsmath}
\usepackage{amsthm}
\usepackage{mathtools}
\usepackage{booktabs}
\usepackage{nicefrac}
\usepackage{multirow}
\usepackage{graphicx}
\usepackage{microtype}
\usepackage{array}
\usepackage{caption}
\usepackage{float}
\usepackage[section]{placeins}
\usepackage{tabularx}
\usepackage{threeparttable}
\usepackage{makecell}
\usepackage{listings}
\usepackage{subcaption}
\usepackage{siunitx}

\theoremstyle{plain}

\theoremstyle{definition}

\theoremstyle{remark}

\lstdefinestyle{feedbackprompt}{
    basicstyle=\ttfamily\footnotesize,
    backgroundcolor=\color{black!3},
    breaklines=true,
    columns=fullflexible,
    frame=single,
    framerule=0.4pt,
    framesep=6pt,
    keepspaces=true,
    rulecolor=\color{black!35},
    xleftmargin=0.04\textwidth,
    xrightmargin=0.04\textwidth,
}

\usepackage{longtable}

\providecommand{\Description}[1]{}
\providecommand{\keywords}[1]{\par\medskip\noindent\textbf{Keywords:} #1\par}

\newenvironment{researchquestions}{%
  \begin{list}{}{%
    \setlength{\leftmargin}{1.5em}%
    \setlength{\labelwidth}{0pt}%
    \setlength{\labelsep}{0pt}%
    \setlength{\itemsep}{3pt}%
    \setlength{\parsep}{0pt}%
    \setlength{\topsep}{3pt}%
  }}{\end{list}}

\begin{document}

\begin{center}
{\Large Robot Programming with Augmented Reality: The Role of Spatial Ability}
\end{center}

\vspace{7mm}

\noindent\textbf{Nicolas Leins}\hfill\href{mailto:leins@zib.de}{\ttfamily leins@zib.de}\\
\emph{\small Zuse Institute Berlin \& TU Berlin, Berlin, Germany}\\
\\
\textbf{Muriel Fischer}\hfill\href{mailto:fischer@zib.de}{\ttfamily fischer@zib.de}\\
\emph{\small Zuse Institute Berlin, Berlin, Germany}\\
\\
\textbf{Malte Teichmann}\hfill\href{mailto:malte.teichmann@wi.uni-potsdam.de}{\ttfamily malte.teichmann@wi.uni-potsdam.de}\\
\emph{\small University of Potsdam \& Weizenbaum Institute Berlin, Germany}\\
\\
\textbf{Jana Gonnermann-M\"uller}\hfill\href{mailto:gonnermann-mueller@zib.de}{\ttfamily gonnermann-mueller@zib.de}\\
\emph{\small Zuse Institute Berlin \& Weizenbaum Institute Berlin, Berlin, Germany}\\
\\
\textbf{Sebastian Pokutta}\hfill\href{mailto:pokutta@zib.de}{\ttfamily pokutta@zib.de}\\
\emph{\small Zuse Institute Berlin \& TU Berlin, Berlin, Germany}\\

\vspace{5mm}

\begin{center}
\begin{minipage}{0.85\textwidth}
\begin{center}
\textbf{Abstract}
\end{center}
{\small
Programming a robot arm requires users to interpret coordinate frames, joint rotations, and trajectories that are not directly visible.
Augmented reality (AR) can make these spatial relations visible, but its benefits may depend on users' spatial ability.
We conducted a randomized between-subjects experiment ($N=71$) in which participants learned to program a physical UR5e robot using either conventional teach-pendant controls with PDF instructions or a head-mounted AR interface that displayed joints, coordinate frames, and waypoints.
We measured users' spatial ability with the Mental Rotation Test and assessed subjective cognitive load and system usability.
Overall, AR did not significantly improve cognitive load or usability compared with conventional instruction.
However, exploratory analyses revealed a compensatory effect: spatial ability predicted higher usability and lower extraneous cognitive load in the control group, but not in the AR condition, suggesting AR mitigated the disadvantage typically faced by users with lower spatial ability. 
These findings point to a compensatory function of AR that should be explored to guide the design of personalized AR interfaces for human-robot interaction.
}
\end{minipage}
\end{center}


\section{Introduction} 
\label{sec1}

Flexible production requires robots to be reprogrammed as tasks change, increasing the need for robot-programming interfaces that are accessible to non-specialists \cite{ivanov_new_2018,chang_survey_2024}. 
Robot programming requires users to connect coordinate frames, joint rotations, and trajectories with movements in the physical workspace. 
Conventional interfaces do not make these connections directly visible, so users must work them out mentally. 
Designing interfaces that support this translation is an important HRI goal \cite{ajaykumar_survey_2021}.

Augmented reality (AR) provides a distinctive form of support. 
By combining spatial tracking with digital overlays, AR devices can place information directly in the physical workspace \cite{azuma_survey_1997}. 
AR robot-programming interfaces have used this capability to display coordinate systems, movement directions, waypoints, and trajectories on or around a robot \cite{yang_har2bot_2024,ikeda_programar_2024}. 
Such in-situ visualizations may reduce the need to mentally translate between instructions and physical robot behavior. 
However, it remains unclear whether these visualizations reduce cognitive demands, are easier for novices to use, or improve on a well-designed conventional system.

Research on AR in HRI has largely focused on developing systems and exploring new interaction techniques \cite{chang_survey_2024}. 
Although some studies examine user experience, small samples and missing control conditions limit conclusions about whether AR offers benefits over conventional interfaces \cite{yang_har2bot_2024,ong_augmented_2020,ikeda_programar_2024,tsamis_intuitive_2021}. 
More controlled comparisons are therefore needed to determine the user experience of AR-supported robot programming.

Reviews and meta-analyses often report benefits of AR for learning, but these benefits vary across studies and designs \cite{howard_meta-analysis_2023,lin_meta-analysis_2023,buchner_media_2023}.
This variation suggests that AR's effects may depend on the task, the form of support, and characteristics of the user.
Yet AR research has paid considerably less attention to learner characteristics than to technological and design factors \cite{gonnermann-muller_unlocking_2025}.
Much of the existing evidence reports only average effects across samples, leaving it unclear whether an interface benefits users with different characteristics equally.

Spatial ability is especially relevant to robot programming because the task requires users to mentally manipulate spatial representations and predict movement in three-dimensional space. 
Two competing accounts describe how spatial ability may affect learning from graphical representations: the ability-as-compensator account predicts greater benefits for learners with lower spatial ability, whereas the ability-as-enhancer account predicts greater benefits for those with higher spatial ability \cite{huk_who_2006}. 
Evidence from AR-supported learning is mixed, and it remains unclear whether either account extends to the dynamic spatial demands of robot programming \cite{kruger_learning_2022,ho_role_2022,bogomolova_effect_2020,ho_effects_2024,weng_effect_2023,kozlova_bringing_2025,gonnermann-muller_unlocking_2025}. 

Clarifying how spatial ability relates to user experience can inform the design of more accessible robot-programming interfaces.
We therefore conducted a randomized between-subjects experiment with $N=71$ participants.
Participants programmed a physical UR5e robotic arm using either teach-pendant controls with PDF instructions or an AR system that reimplemented the core programming controls and added spatially registered visualizations.
We examine two aspects of user experience: system usability and cognitive load. 
System usability captures participants' evaluation of the complete instructional and interaction system, whereas cognitive load captures the mental demands they experienced during the task \cite{vlachogianni_perceived_2022,ardito_approach_2006,sweller_cognitive_2019}.

The study addresses three research questions:

\begin{researchquestions}
    \item \textbf{RQ1}. To what extent does AR improve user experience during a robot-programming task compared with the control condition?
    \item \textbf{RQ2}. How is spatial ability associated with user experience during robot programming?
    \item \textbf{RQ3}. Does the relationship between spatial ability and user experience differ between the AR and control conditions?
\end{researchquestions}

\textbf{The paper makes three contributions:}
\begin{itemize}
    \item a controlled comparison of AR-supported and conventional programming of a physical robot, which found no measurable group-level advantage of AR in cognitive load or system usability;
    \item evidence that spatial ability is associated with user experience during robot programming, with higher spatial ability related to lower extraneous cognitive load and higher system usability; and
    \item exploratory condition-specific results showing that these associations were present in the conventional condition but not in the AR condition, a pattern consistent with AR compensating for lower spatial ability.
\end{itemize}

\section{Theoretical background}
\label{theoretical-background}

\subsection{Augmented reality for robot programming}
\label{learning-with-AR}
\label{AR-for-HRI}

Within HRI, AR is increasingly used to support teleoperation, pick-and-place tasks, and industrial robot programming \cite{chang_survey_2024,hedayati_improving_2018,walker_communicating_2018}.
Robot programming is a particularly relevant application for AR, as users must connect interface representations to positions, orientations, and movements in the physical workspace.
AR can support this connection by overlaying spatially registered virtual information onto the real environment through handheld displays, head-mounted displays, or spatial AR \cite{azuma_survey_1997,carmigniani_augmented_2011}.

From a user perspective, the three core technological characteristics of AR identified by \citet{azuma_survey_1997} can be reframed as \textit{Contextuality}, \textit{Interactivity}, and \textit{Spatiality} \cite{kruger_augmented_2019}.
\textit{Contextuality} refers to embedding virtual elements in the real environment, grounding information where it is needed and supporting spatial and temporal contiguity \cite{sautter_mixed_2021,ayres_split-attention_2014,mayer_multimedia_2020}.
\textit{Interactivity} captures the real-time, embodied responsiveness of virtual elements to user input, such as touching, rotating, or moving around objects \cite{holmes_move_2018,johnson-glenberg_embodied_2017}.
\textit{Spatiality} refers to accurately placing and aligning virtual objects in physical space \cite{azuma_survey_1997,jerabek_perceptual_2015}, which is valuable for tasks involving positions and spatial relationships \cite{radu_augmented_2014}.

These characteristics map directly onto the demands of robot programming.
\textit{Spatiality} can externalize invisible coordinate frames and movement consequences; \textit{Contextuality} can place instructions alongside the physical robot; and \textit{Interactivity} can allow users to inspect programmed positions and movements while working with the robot.
Together, these characteristics may reduce the mental translation between interface instructions and physical robot behavior.

Existing AR systems for HRI mark safety zones and task-relevant positions \cite{ikeda_programar_2024,tsamis_intuitive_2021,makris_augmented_2016}, preview planned movements and trajectories \cite{ikeda_programar_2024,ikeda_marcer_2025,lotsaris_ar_2021,yang_har2bot_2024,ong_augmented_2020}, visualize waypoints \cite{neves_application_2020,yang_har2bot_2024}, and support gesture-based definition of positions in the workspace \cite{lotsaris_ar_2021,ong_augmented_2020,tsamis_intuitive_2021,yang_har2bot_2024,jiang_comprehensive_2024}.
Prior studies associate these features with faster programming, higher success rates, lower cognitive load, improved usability, and enhanced safety \cite{ong_augmented_2020,tsamis_intuitive_2021,yang_har2bot_2024}, as well as reduced task complexity for non-experts \cite{ikeda_programar_2024,lotsaris_ar_2021,neves_application_2020,ong_augmented_2020,yang_har2bot_2024,moorman_teaching_2026}.
However, the evidence base remains limited because studies commonly use small samples \cite{yang_har2bot_2024,ong_augmented_2020,ikeda_programar_2024,tsamis_intuitive_2021}, lack control groups \cite{ikeda_programar_2024}, rely on descriptive designs \cite{ong_augmented_2020}, or omit evaluation \cite{lotsaris_ar_2021,neves_application_2020,makris_augmented_2016}.

Research beyond HRI shows that AR effects vary across studies and designs \cite{howard_meta-analysis_2023,lin_meta-analysis_2023,buchner_media_2023}.
Media comparisons alone therefore cannot explain when AR provides useful support \cite{buchner_media_2023,lee_augmented_2012,webel_augmented_2013}.
Controlled comparisons are needed to establish whether AR improves user experience in robot programming, but average differences may still conceal variation between users.
This motivates examining spatial ability as a user characteristic which may help explain differences in user experience with AR \cite{kozlova_bringing_2025,gonnermann-muller_unlocking_2025}.

\subsection{Spatial ability in AR-supported robot programming}

Spatial abilities enable individuals to visualize, mentally manipulate, and reason about objects in three-dimensional space \cite{carroll_human_1993}.
Rather than a single, unified skill, spatial ability is multidimensional, with spatial perception, mental rotation, and spatial visualization commonly distinguished \cite{linn_emergence_1985}.
This study focuses on mental rotation, the capacity to mentally rotate objects without physical aids.
In the following, spatial ability therefore refers specifically to mental rotation.

Robot programming places direct demands on spatial reasoning because users must anticipate how joint rotations, coordinate transformations, and programmed trajectories will affect physical robot movement.
Users with higher spatial ability may be better able to construct these representations mentally, which could reduce unnecessary cognitive demands and improve their experience of the programming system.

Two competing hypotheses describe how spatial ability may relate to learning from graphical representations \cite{huk_who_2006}.
Under the ability-as-enhancer hypothesis, learners with higher spatial ability benefit more because they have more cognitive resources available to process and integrate visual information.
Under the ability-as-compensator hypothesis, learners with lower spatial ability benefit more because graphical representations reduce the need to construct spatial information mentally.
Although developed for graphical representations more broadly, both hypotheses can plausibly be applied to AR-supported robot programming.
If AR visualizations require substantial spatial processing, users with higher spatial ability may benefit more; if they externalize relations that users would otherwise construct mentally, users with lower spatial ability may benefit more.

Empirical studies of AR-supported learning provide mixed support for these hypotheses.
Studies using AR or mixed-reality anatomy visualizations have reported patterns consistent with the ability-as-enhancer hypothesis \cite{kruger_learning_2022,ho_role_2022}.
Other studies have reported patterns consistent with the ability-as-compensator hypothesis in anatomy learning and construction training \cite{bogomolova_effect_2020,ho_effects_2024}, or found no effect of spatial ability on learning outcomes or satisfaction \cite{weng_effect_2023}.

Much of this evidence concerns relatively static learning content, particularly anatomy visualizations \cite{bogomolova_effect_2020,ho_role_2022,kruger_learning_2022}.
Robot programming, by contrast, requires users to reason about changing spatial relations while controlling a physical robot.
It therefore remains unclear which, if any, of these accounts found in AR-supported learning apply to user experience with AR-supported robot programming \cite{kozlova_bringing_2025,gonnermann-muller_unlocking_2025}.

\subsection{Synthesis and hypotheses}
\label{Synthesis}

Robot-programming interfaces differ in how much spatial translation they require users to perform mentally.
Conventional screen-based instructions require users to relate interface representations to movements in the physical workspace.
By visualizing joint rotations, coordinate systems, and waypoints directly on or around the robot, AR may externalize some of these relations.
This support primarily draws on AR's \textit{Spatiality} and \textit{Contextuality} and may reduce the need to divide attention between instructions, controls, and robot movement \cite{ayres_split-attention_2014,mayer_multimedia_2020,lin_meta-analysis_2023}.

We examine how this support and users' spatial ability relate to two aspects of user experience: system usability and cognitive load.
System usability captures how participants evaluate the complete instructional and interaction system.
Poor usability can direct effort toward operating the technology rather than completing the task \cite{vlachogianni_perceived_2022,ardito_approach_2006}.

Cognitive load captures the working-memory demands users experience during a task \cite{sweller_cognitive_2019}.
It can arise from intrinsic cognitive load (ICL), determined by task complexity and prior knowledge; extraneous cognitive load (ECL), imposed by factors such as instructional and interface design; and germane cognitive load (GCL), associated with integrating new information \cite{moreno_cognitive_2010,chandler_cognitive_1996,sweller_cognitive_1998,sweller_cognitive_2019,sweller_element_2010}.
ECL is central to the conceptual framework because it captures demands created by the interface and instructional design rather than by the robot-programming task itself.
We report ICL and GCL to characterize participants' cognitive load but do not formulate directional hypotheses for these dimensions.

Against this background, RQ1 examines whether AR improves user experience compared with conventional instruction, leading to the following hypotheses.

    \begin{itemize}
        \item \textbf{H1a}. The AR condition reduces extraneous cognitive load compared with the control condition.
        \item \textbf{H1b}. The AR condition increases system usability compared with the control condition.
    \end{itemize}

Independent of the interface, users with higher spatial ability should be better able to interpret the spatial representations required for robot programming.
They should therefore experience lower ECL and higher system usability.
RQ2, examining how spatial ability relates to user experience, leads to the following hypotheses.

    \begin{itemize}
        \item \textbf{H2a}. Higher spatial ability is negatively associated with extraneous cognitive load.
        \item \textbf{H2b}. Higher spatial ability is positively associated with system usability.
    \end{itemize}

RQ3 asks whether these associations differ between the AR and control conditions.
Under the ability-as-compensator account, AR should reduce the dependence of user experience on spatial ability by supporting users with lower spatial ability.
Under the ability-as-enhancer account, the association should become stronger because users with higher spatial ability can make better use of the visualizations.
Because these accounts make competing predictions, we examine RQ3 exploratorily without a directional hypothesis.

\section{Method}
\label{Method}

\subsection{Participants}
\label{Participants}

Seventy-four individuals initially took part in the experiment.
Three were excluded before analysis (one for an incorrectly completed questionnaire; two required substantial help and could not complete the task independently).
The final sample consisted of $N=71$ participants (19 female and 52 male, $M_{age}=26.58$ years, $SD=6.90$ years).
Participants were randomly allocated to conditions, aiming for an equal distribution of sexes across conditions to minimize sex-related differences in spatial ability \cite{castro-alonso_sex_2019,peters_redrawn_1995}.
Of the final sample, the AR condition included $n=36$ participants (9 female, 27 male; $M_{age}=26.92$, $SD=6.69$), and the control condition included $n=35$ participants (10 female, 25 male; $M_{age}=26.23$, $SD=7.20$).

\subsection{Procedure and experimental design}
\label{Procedure}

We conducted a randomized between-subjects experiment (approved by the ethics committee of the authors' research institute), comparing the control condition (conventional interface with PDF instructions on a tablet) with the AR condition (equivalent controls and instructions, supplemented with AR-specific learning materials); both used the same robot programming task, differing only in instructional method, with a between-subjects design chosen to eliminate potential learning-transfer effects.

All sessions took place in an industrial laboratory environment equipped with a physical 6-axis robotic arm (\autoref{fig:experiment-setup}).
The combination of a physical industrial robot and a production-oriented pick-and-place task provided an ecologically meaningful HRI setting, while the laboratory allowed us to control the environment and procedure.
Bright, standardized lighting ensured reliable AR tracking and video-see-through quality.
The procedure followed a standardized protocol: after providing informed consent, participants received an introduction to the task, were encouraged to work independently, and were told no prior knowledge was required.
We then explained device and interface handling, and AR participants received a demo application on the AR headset to build familiarity.
After receiving safety information (safety areas, reach limits, emergency stop procedures), participants worked independently through the step-by-step learning task until the final program executed correctly, concluding with a final questionnaire.

\subsection{Robot-programming task}
\label{Learning-task}

The robot-programming task was designed to require spatial reasoning while remaining solvable without prior robot-programming experience.

\begin{figure}[ht]
    \centering
    \includegraphics[width=0.7\linewidth]{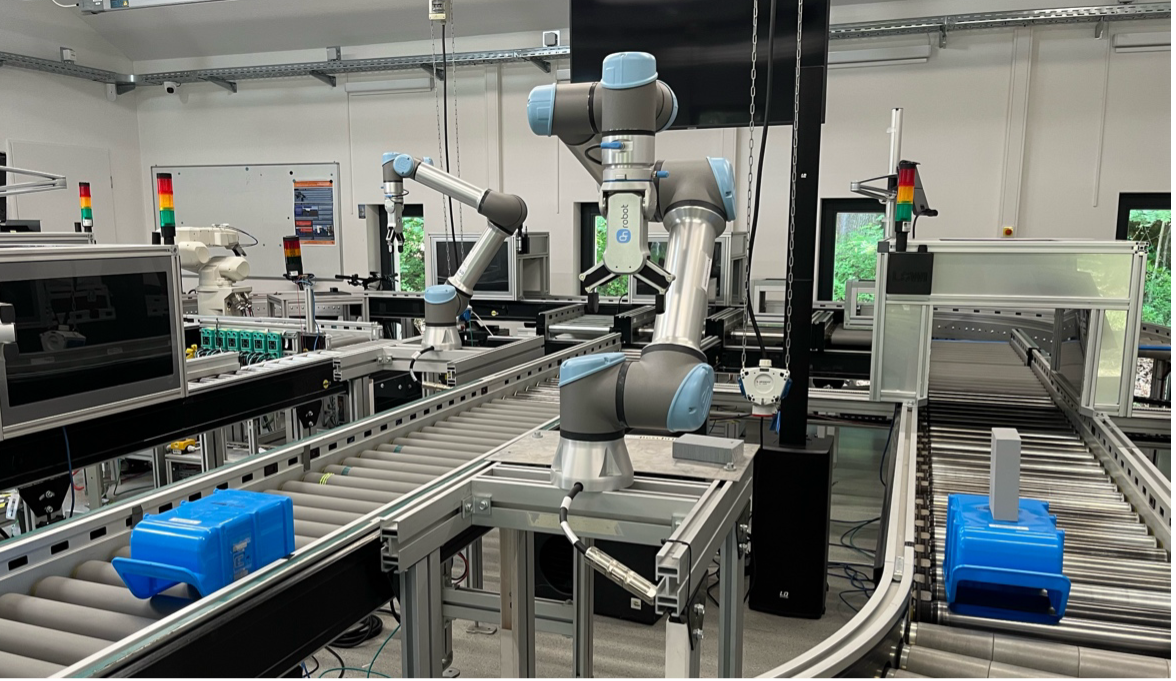}
    \caption{Experimental setup: UR5e arm with RG2 gripper, workpiece (grey cube), pickup (right) and target (left) carriers.}
    \label{fig:experiment-setup}
    \Description{A UR5e robot arm with an RG2 gripper stands between two conveyor lines in an industrial laboratory.
    A gray workpiece rests on a blue carrier on the right, and an empty blue target carrier is positioned on the left.}
\end{figure}

Participants completed the task using a UR5e from Universal Robots, fitted with an OnRobot RG2 gripper.
The task progressed from joint rotation control to translational tool-center-point (TCP) movement, to programming movements and gripper actions with collision-avoiding support positions. 
Participants concluded by programming a pick-and-place sequence: picking up a workpiece and placing it on a second carrier (see \autoref{fig:experiment-setup} for the experimental setup) and modifying the program as needed until it executed correctly. 

\subsection{Instruction material}
\label{learning-material}

We developed dedicated instruction materials for each condition. 
The task and content were equal across conditions, with only the user interface differing due to the technical device and additional AR visualizations.
The following sections describe the conventional (PDF) material, the AR application, and the pretest used to validate both before the main study. 

\subsubsection{Control condition}
\label{conventional-learningmaterial}

\begin{figure}[ht]
    \centering
    \begin{subfigure}[t]{0.48\linewidth}
        \centering
        \includegraphics[width=\linewidth]{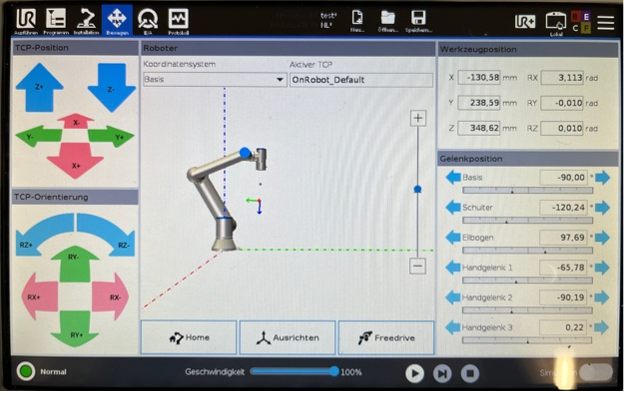}
        \caption{}
        \label{fig:teach-pendant-movement}
    \end{subfigure}
    \hspace{0.02\linewidth}
    \begin{subfigure}[t]{0.48\linewidth}
        \centering
        \includegraphics[width=\linewidth]{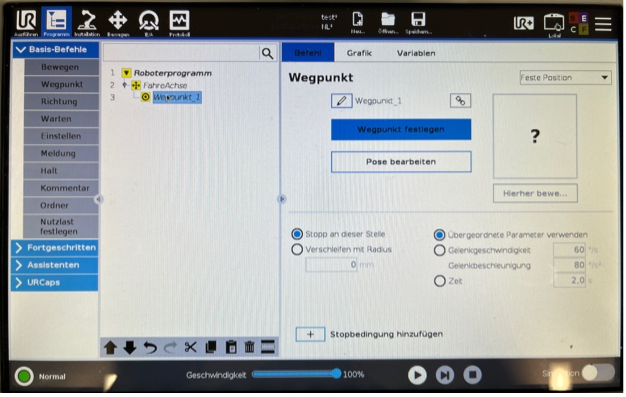}
        \caption{}
        \label{fig:teach-pendant-program}
    \end{subfigure}
    \caption{Teach pendant user interface.
    Movement tab (a) and program tab (b).}
    \Description{Two teach-pendant screens.
    Panel (a) shows controls for TCP and joint movement around a robot model.
    Panel (b) shows the program tree and waypoint controls.}
    \label{fig:teach-pendant-interface}
\end{figure}

Participants in the control condition used UR5e's proprietary teach pendant running PolyScope 5.0 \cite{noauthor_polyscope_2026}, using its movement tab for joint rotation and TCP control (\autoref{fig:teach-pendant-movement}) and its program tab for the programming task (\autoref{fig:teach-pendant-program}).

Instructions were delivered via PDF on an iPad 10.2.
Participants swiped and zoomed through the material as needed.
We placed both devices on a standing desk in front of the robot. 

\subsubsection{Augmented reality application}
\label{AR-application}

\begin{figure}[ht]
    \centering
    \includegraphics[width=0.8\linewidth]{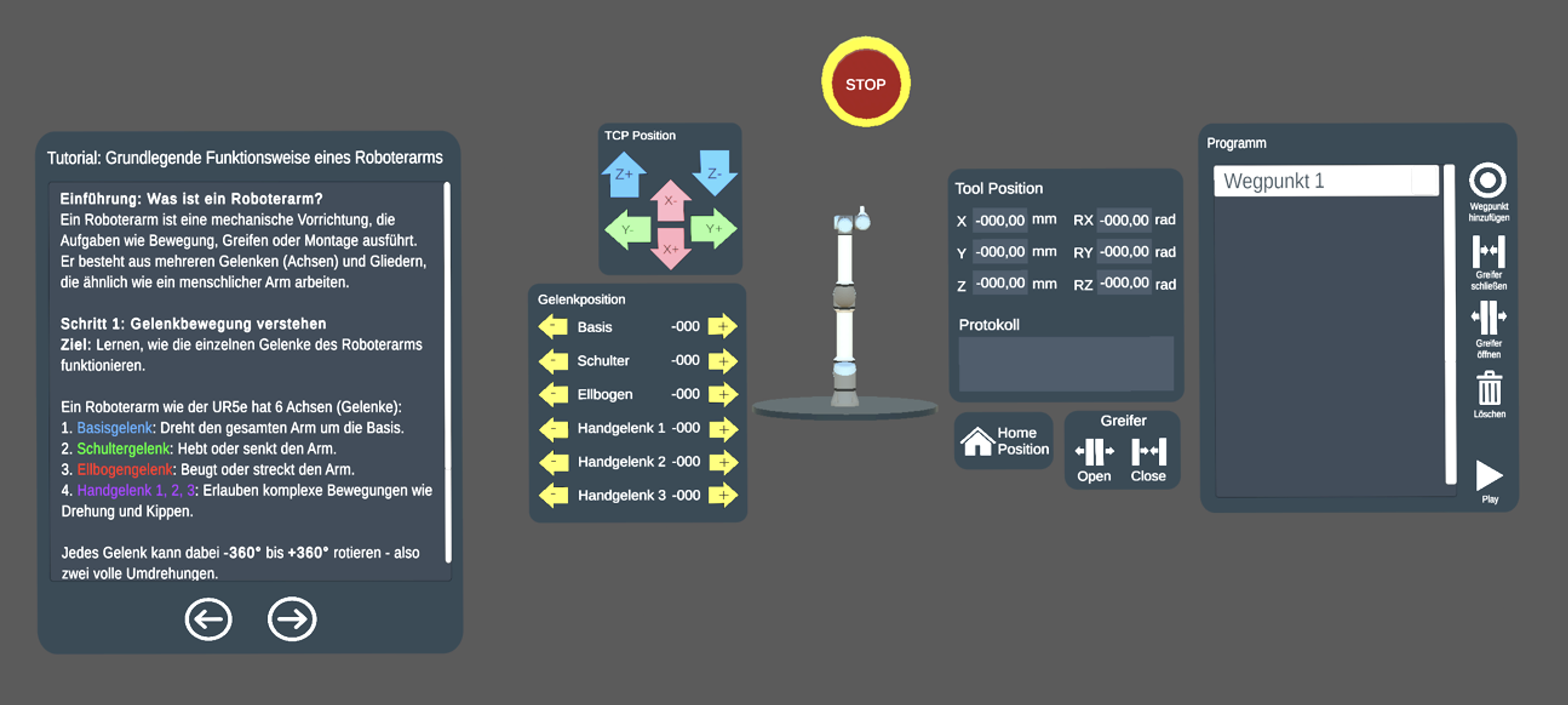}
    \caption{User interface of the AR application, showing instruction, movement control, emergency stop, tool/position information, and program panels.}
    \Description{The AR interface contains a tutorial panel, TCP and joint controls, an emergency-stop button, a robot and tool-position display, home-position and gripper controls, and a waypoint-programming panel.}
    \label{fig:ar-interface}
\end{figure}

We developed a corresponding AR application for the Meta Quest 3 (video-see-through HMD AR) using Unity 2022.3.
The complete Unity project for the AR application, including all source code and scripts, will be made publicly available upon acceptance.
As small on-device text is difficult to perceive through the Quest 3's video-see-through cameras (18 PPD) despite its wide field of view (110°×96°), all controls and information were natively reimplemented in AR (\autoref{fig:ar-interface}), with joint-based and translational TCP movement and hand-tracked, touchscreen-style button interaction.
To preserve comparability with the control condition, the application replicated the teach pendant's core functionality and interface logic (joint/translational movement, waypoint programming) rather than redesigning the interaction paradigm, as done in prior work \cite{neves_application_2020,ong_augmented_2020,yang_har2bot_2024}.

\begin{figure}[!ht]
    \centering
    \begin{minipage}{0.8\linewidth}
    \newlength{\arfigureheight}
    \setlength{\arfigureheight}{0.39\linewidth}
    \begin{subfigure}[t]{0.29\linewidth}
        \centering
        \includegraphics[height=\arfigureheight]{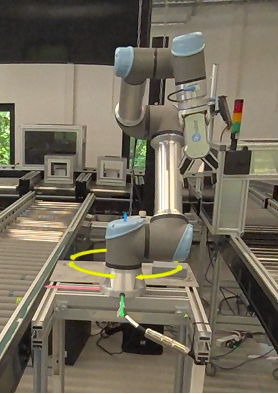}
        \caption{}
        \label{fig:base-rotation-visualization}
    \end{subfigure}\hfill%
    \begin{subfigure}[t]{0.29\linewidth}
        \centering
        \includegraphics[height=\arfigureheight]{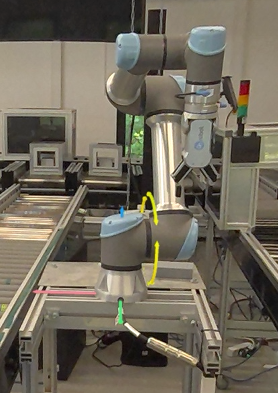}
        \caption{}
        \label{fig:shoulder-rotation-visualization}
    \end{subfigure}\hfill%
    \begin{subfigure}[t]{0.36\linewidth}
        \centering
        \includegraphics[height=\arfigureheight]{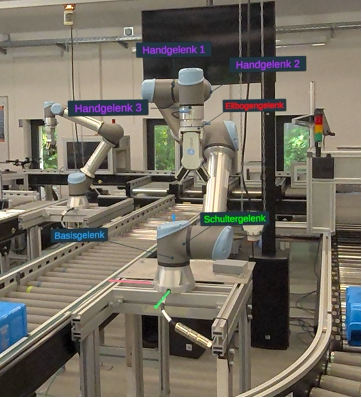}
        \caption{}
        \label{fig:joint-label}
    \end{subfigure}

    \vspace{0.5em}
    \begin{subfigure}[t]{0.58\linewidth}
        \centering
        \includegraphics[height=\arfigureheight]{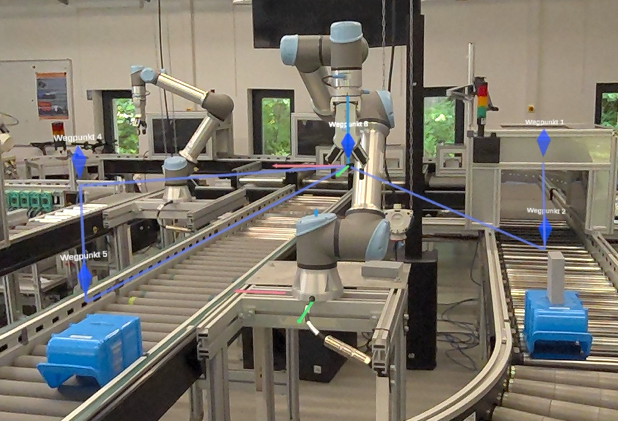}
        \caption{}
        \label{fig:waypoints}
    \end{subfigure}\hfill%
    \begin{subfigure}[t]{0.36\linewidth}
        \centering
        \includegraphics[height=\arfigureheight]{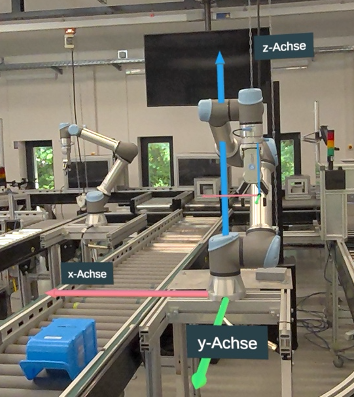}
        \caption{}
        \label{fig:base-coordinate-system}
    \end{subfigure}
    \end{minipage}
    \caption{AR visualizations: rotation of the base joint (a) and shoulder joint (b); joint labels (c); coordinate system at the TCP and waypoint and path visualization (d); and base coordinate system at the robot base (e).}
    \Description{Five AR views of the robot.
    Panels (a) and (b) overlay yellow arrows indicating base and shoulder rotation.
    Panel (c) labels the robot's joints.
    Panel (d) shows the TCP coordinate system and blue waypoint markers connected by the programmed path.
    Panel (e) overlays the x-, y-, and z-axes at the robot base.}
    \label{fig:ar-visualizations}
\end{figure}

The AR device communicated with the robot via ROS2 using Universal Robots' ROS2 driver \cite{universal_robots_as_universal_2025} and Unity ROS2 package \cite{unity_technologies_ros_2025}, while URScript commands controlled robot motion.
The Unity application continuously received the robot's joint positions to mirror its live pose.
A manually defined spatial anchor at the robot base aligned virtual content with the physical robot, enabling three AR-specific features used as the experimental manipulation:

First, rotation arrows were dynamically rendered at the active joint during joint movement (\autoref{fig:base-rotation-visualization} and \ref{fig:shoulder-rotation-visualization}), showing the axis and direction at the joint itself to reduce the need to mentally simulate joint motion.
Second, we visualized an offset world coordinate system directly at the TCP during translational movement (\autoref{fig:waypoints}), allowing the correct movement axis to be identified perceptually rather than through mental spatial transformation.
Third, we visualized programmed waypoints as labeled, connected markers at their real-world positions (\autoref{fig:waypoints}), externalizing the programmed path as a spatial structure that could be visually traced and checked for errors before execution, rather than reconstructed from numeric values on a screen.
We also reimplemented the static picture-based visualizations from the PDF instructions (joint labels and base coordinate system) directly on the real robot, shown and hidden dynamically per instructional step (\autoref{fig:joint-label} and \ref{fig:base-coordinate-system}).

These features map onto the three AR characteristics introduced in the theoretical background \cite{kruger_augmented_2019}: joint rotation arrows and coordinate-system visualizations primarily instantiate \textit{Spatiality} and \textit{Contextuality}, while waypoint and path visualization combine all three, situating poses in the workspace (\textit{Contextuality}), aligning them in 3D space (\textit{Spatiality}), and supporting interactive programming (\textit{Interactivity}). 
Throughout, the design preserved learner agency via full manual control, natural hand input, and self-paced instruction, so that AR scaffolded spatial reasoning without automating task execution.

\subsubsection{Pretest}

We pretested the task and both materials with six participants (three per condition) using the think-aloud method \cite{ericsson_verbal_1980}. 
Both materials proved sufficiently designed (completion times: $16-28$ minutes), and feedback led to refinements improving clarity and cross-condition consistency, including consistently presenting instructions before each task, highlighting relevant UI elements, condensing AR instructions to single non-scrolling screens, and revising the control condition's gripper-programming instructions to match participants' more intuitive approach.

\subsection{Measurements}
\label{Measurements}

A questionnaire collected participants' age and sex, followed by measures of user experience, spatial ability, technology affinity, and prior experience with AR, robotics, and programming.
Technology affinity and prior experience were assessed as background variables to characterize the sample and examine group comparability; they were not included as predictors in the main analyses.
Instrument details are summarized in \autoref{tab:instruments}.
In the questionnaire, we defined the "system" to evaluate per condition as the full interaction system (AR condition: device, application, and instructional content; control condition: tablet PDF manual and teach pendant), ensuring participants evaluated their complete experience rather than isolated components.

\begin{table}[ht]
    \centering
    \caption{Measurement instruments.}
    \label{tab:instruments}
    \resizebox{\columnwidth}{!}{%
    \begin{tabular}{l l l l}
        \toprule
        \textbf{Construct} & \textbf{Instrument} & \textbf{Items / Scale} & \textbf{Reliability} \\
        \midrule
        Usability & SUS \cite{brooke_sus_1996,gao_multi-language_2020} & 10 items, 5-point & $\alpha = .83$ \\
        \addlinespace
        \multirow{3}{*}{Cognitive load} & \multirow{3}{*}{\citet{klepsch_development_2017}} & \multirow{3}{*}{3 subscales (ECL, ICL, GCL), 7-point} & ECL: $\alpha = .71$ \\
         & & & ICL: $r_{SB} = .78$ \\
         & & & GCL: $\alpha = .31$ \\
        \addlinespace
        Spatial ability & MRT \cite{peters_redrawn_1995} & 20 items (2 blocks of 10), 0--20 score & --- \\
        \addlinespace
        Technology affinity & ATI \cite{franke_personal_2018} & 9 items, 6-point & $\alpha = .93$ \\
        \addlinespace
        \multirow{3}{*}{Prior experience} & AR & \multirow{3}{*}{Single item each, 5-point} & \multirow{3}{*}{---} \\
         & Robotics & & \\
         & Programming & & \\
        \bottomrule
    \end{tabular}
    }
    \vspace{1ex}
    
    \raggedright
    \footnotesize
    Note.
    SUS = System Usability Scale; MRT = Mental Rotation Test; ATI = Affinity for Technology Interaction; ECL/ICL/GCL = extraneous/intrinsic/germane cognitive load; $r_{SB}$ = Spearman-Brown coefficient.
    All instruments were administered in German.
\end{table}

\subsection{Data analysis}
\label{Data-analysis}

All analyses were conducted in R.
Anonymized study data and R analysis scripts are available separately in an \href{https://osf.io/4jezs/overview?view_only=c2625cb7183449f4a8e93fbd062abfac}{OSF repository}.

We examined condition (AR vs.\ control) and spatial ability (MRT score) as predictors of user experience.
Instructional design was also examined exploratorily as a moderator of the association between spatial ability and user experience (RQ3).
System usability and cognitive load served as dependent measures of user experience.

Group equivalence on learner characteristics (MRT, ATI, and prior experience with AR (EAR), robotics (ER), and programming (EP)) was tested using Welch's t-test (normally distributed variables) or the Mann-Whitney-U test (non-normally distributed variables). 

RQ1 and RQ2 were addressed via four multiple linear regressions (DVs: ECL, ICL, GCL, SUS) with condition and MRT entered simultaneously as predictors.
RQ3 was addressed via subgroup analyses, estimating separate regressions with MRT as a predictor within each condition and comparing coefficients across conditions.

Model assumptions were evaluated via Shapiro-Wilk tests and Q-Q plots (normality), Levene's tests (variance homogeneity), and residual plots and Breusch-Pagan tests (homoscedasticity). 
Raw scores and residuals departed from normality for all outcomes ($p < .05$), common for bounded Likert-type scales, but Levene's tests showed no significant between-group variance differences (all $p \geq .05$), and heteroscedasticity was detected only for ECL ($p < .001$; all others $p > .07$). 
Given balanced group sizes, the absence of meaningful variance differences, and OLS's robustness under moderate distributional violations, parametric models were retained for all outcomes, with heteroscedasticity-consistent (HC3) standard errors applied to ECL models (including RQ3 subgroup regressions) and conventional standard errors otherwise.
Within each predictor, coefficient $p$-values were Bonferroni-corrected across the four outcome models ($\times 4$, $\alpha=.05$); omnibus model $p$-values and 95\% CIs were unadjusted.
Power analyses were additionally conducted for both predictors (sensitivity analysis for the condition effect; achieved power for the MRT effect).

\section{Results}
\label{Results}

To address RQ1, we examined how the AR condition influenced user experience during a robot-programming task compared with the control condition.
RQ2 focused on the role of participants' spatial ability in shaping their user experience during robot programming.
Using an exploratory approach, RQ3 examined whether AR moderated the relationship between spatial ability and user experience in a manner consistent with compensation or enhancement.

\autoref{tab:descriptives} reports descriptive statistics for task duration and all outcome variables, overall and by condition.
On average, participants required $M=24.87$ minutes to complete the task, with participants in the AR condition finishing faster than those in the control condition.
The task was rated with low-to-medium ICL and ECL and high GCL, and usability (SUS) was rated favorably in both conditions.

\begin{table}[ht]
    \centering
    \caption{Descriptive statistics for task duration and outcome variables.}
    \label{tab:descriptives}
    \resizebox{0.66\columnwidth}{!}{%
    \begin{tabular}{l c c c}
        \toprule
        \textbf{Variable} & \textbf{Full sample} & \textbf{AR} ($n=36$) & \textbf{Control} ($n=35$) \\
        & $M$ ($SD$) & $M$ ($SD$) & $M$ ($SD$) \\
        \midrule
        Duration (min) & 24.87 (5.73) & 23.11 (4.82) & 26.69 (6.08) \\
        \addlinespace
        ICL & 2.76 (1.22) & 2.40 (0.94) & 3.13 (1.37) \\
        ECL & 1.79 (0.83) & 1.70 (0.70) & 1.88 (0.95) \\
        GCL & 5.66 (0.91) & 5.61 (0.85) & 5.71 (0.97) \\
        SUS & 80.95 (14.08) & 82.57 (14.10) & 79.29 (14.07) \\
        \bottomrule
    \end{tabular}
    }
    \vspace{1ex}
    
    \raggedright
    \footnotesize
    Note.
    ICL = intrinsic cognitive load; ECL = extraneous cognitive load; GCL = germane cognitive load; SUS = System Usability Scale.
\end{table}

Inferential statistical tests confirmed the comparability of the groups on MRT, ATI, and prior experience (\autoref{tab:demographics}). 
MRT scores did not differ significantly between groups, $t(68.36) = -0.50$, $p = .621$, 95\% CI $[-2.55, 1.54]$.

\begin{table}[ht]
    \centering
    \caption{Group equivalence on spatial ability, technology affinity, and prior experience.}
    \label{tab:demographics}
    \resizebox{0.66\columnwidth}{!}{%
    \begin{tabular}{l
                    S[table-format=2.2]
                    S[table-format=1.2]
                    S[table-format=2.2]
                    S[table-format=1.2]
                    l
                    S[table-format=1.3]}
        \toprule
        & \multicolumn{2}{c}{\textbf{AR} ($n=36$)} & \multicolumn{2}{c}{\textbf{Control} ($n=35$)} & & \\
        \textbf{Variable} & {$M$} & {$SD$} & {$M$} & {$SD$} & \textbf{Test statistic} & {$p$} \\
        \midrule
        MRT & 10.22 & 4.16 & 9.71 & 4.46 & $t(68.36)=-0.50$ & .621 \\
        ATI & 4.28  & 1.05 & 4.46 & 1.09 & $U=702$   & .410 \\
        ER  & 1.58  & 0.94 & 1.57 & 1.09 & $U=586$   & .540 \\
        EP  & 3.11  & 1.30 & 2.89 & 1.39 & $U=572.5$ & .499 \\
        EAR & 2.14  & 1.15 & 2.23 & 1.11 & $U=672$   & .618 \\
        \bottomrule
    \end{tabular}
    }
    \vspace{1ex}
    
    \raggedright
    \footnotesize
    Note.
    MRT = Mental Rotation Test; ATI = Affinity for Technology Interaction; ER/EP/EAR = prior experience with robotics/programming/AR.
    MRT was tested with Welch's $t$-test (95\% CI $[-2.55, 1.54]$); others were tested with Mann-Whitney $U$ ($N_1=35$, $N_2=36$).
    No condition differences were significant.
\end{table}

Male participants scored significantly higher on MRT than female participants ($M_{\mathrm{male}}=11.12$, $SD=3.90$ vs.\ $M_{\mathrm{female}}=6.84$, $SD=3.78$), $t(32.997)=4.185$, $p<.001$, 95\% CI $[2.20, 6.35]$, Cohen's $d=1.105$ (large effect). 
However, because overall MRT did not differ between groups and sex was evenly distributed across conditions, we did not examine these effects further.

\subsection{Effects of instructional condition}
\label{results-RQ1}

To address RQ1, we conducted multiple linear regression models with condition and MRT score as predictors (\autoref{tab:rq1rq2-regression}).
The effects of instructional condition on ECL and SUS tested H1a and H1b, while ICL and GCL were included to characterize participants' cognitive load.
Based on the omnibus tests, regression models predicting ECL, ICL, and SUS were significant (all $p<.01$), while the model for GCL was not ($p=.34$); the GCL scale's low internal consistency (Cronbach's $\alpha=.31$) may limit the reliability of this null result.
Across all four models, the condition predictor was not significant.
Thus, neither H1a nor H1b was supported, and instructional condition also did not significantly predict ICL or GCL.

A sensitivity power analysis ($N=71$; $df=1,68$) confirmed adequate power to detect medium-to-large effects of instructional condition (at least $.89$ at $\alpha=.05$ and at least $.75$, Bonferroni-adjusted), though power for small effects ($f^2=.02$) was limited ($\leq.22$); the non-significant condition effects thus likely reflect a true absence of medium-to-large differences.

\begin{table}[ht]
    \centering
    \caption{Regression results for RQ1 and RQ2: instructional condition and spatial ability (MRT) predicting user-experience outcomes.}
    \label{tab:rq1rq2-regression}
    \resizebox{\columnwidth}{!}{%
    \begin{tabular}{l l l l l
                    S[table-format=-1.2]
                    S[table-format=1.2]
                    S[table-format=-1.2]
                    l
                    l}
        \toprule
        \textbf{Outcome} & {$F(2,68)$} & {$p_{\mathrm{model}}$} & {$R^2$} & \textbf{Predictor} & {$b$} & {$SE$} & {$\beta$} & {$p_{\mathrm{adj}}$} & {95\% CI} \\
        \midrule
        \multirow{2}{*}{ECL} & \multirow{2}{*}{5.65} & \multirow{2}{*}{<.01} & \multirow{2}{*}{.14} & Condition & -0.14 & 0.19 & -0.08 & 1.00 & [-0.51, 0.24] \\
        & & & & MRT & -0.07 & 0.02 & -0.36 & <.01 & [-0.115, -0.026] \\
        \addlinespace
        \multirow{2}{*}{ICL} & \multirow{2}{*}{5.05} & \multirow{2}{*}{<.01} & \multirow{2}{*}{.13} & Condition & -0.70 & 0.27 & -0.29 & .053 & [-1.24, -0.15] \\
        & & & & MRT & -0.06 & 0.03 & -0.20 & .34 & [-0.121, 0.008] \\
        \addlinespace
        \multirow{2}{*}{GCL} & \multirow{2}{*}{1.09} & \multirow{2}{*}{.34} & \multirow{2}{*}{.03} & Condition & -0.12 & 0.22 & -0.07 & 1.00 & [-0.55, 0.31] \\
        & & & & MRT & 0.04 & 0.03 & 0.17 & .67 & [-0.015, 0.086] \\
        \addlinespace
        \multirow{2}{*}{SUS} & \multirow{2}{*}{6.07} & \multirow{2}{*}{<.01} & \multirow{2}{*}{.15} & Condition & 2.66 & 3.13 & 0.10 & 1.00 & [-3.58, 8.91] \\
        & & & & MRT & 1.22 & 0.37 & 0.37 & <.01 & [0.488, 1.955] \\
        \bottomrule
    \end{tabular}
    }
    \vspace{1ex}
    
    \raggedright
    \footnotesize
    Note.
    ECL/ICL/GCL = extraneous/intrinsic/germane cognitive load; SUS = System Usability Scale; Condition = experimental condition (AR vs.\ control).
    Omnibus model $p$-values and 95\% CIs are unadjusted.
    Predictor $p$-values are Bonferroni-adjusted separately for condition and MRT across the four outcomes ($\times4$).
\end{table}

\subsection{Associations with spatial ability}
\label{results-RQ2}

\begin{figure}[ht]
    \centering
    \begin{subfigure}[t]{0.45\linewidth}
        \centering
        \includegraphics[width=\linewidth]{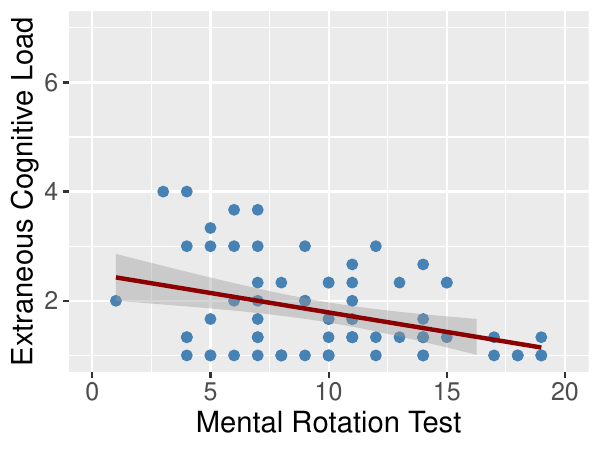}
        \caption{}
        \label{fig:ecl-mrt}
    \end{subfigure}\hfill
    \begin{subfigure}[t]{0.45\linewidth}
        \centering
        \includegraphics[width=\linewidth]{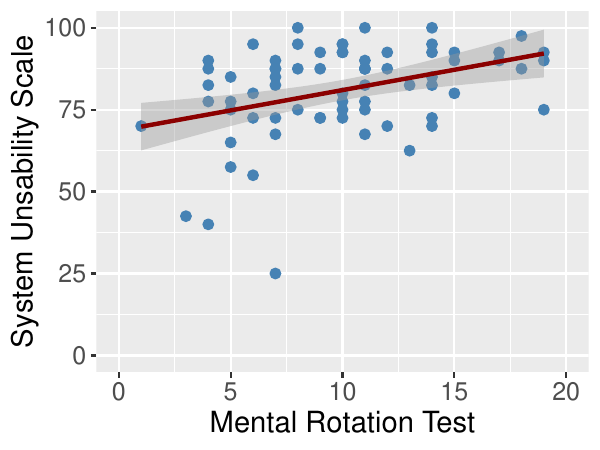}
        \caption{}
        \label{fig:sus-mrt}
    \end{subfigure}
    \caption{ECL (a) and system usability (b) as a function of MRT score, with regression line and 95\% CI.}
    \Description{Two scatterplots with regression lines and confidence bands.
    Extraneous cognitive load decreases as Mental Rotation Test score increases, while system usability increases as Mental Rotation Test score increases.}
    \label{fig:sus-ecl-mrt}
\end{figure}

To examine the influence of spatial ability on user experience (RQ2), we analyzed the MRT predictor within the same regression models (\autoref{tab:rq1rq2-regression}).
MRT score significantly predicted ECL and SUS (both $p<.01$), with higher spatial ability associated with lower ECL and higher usability (see \autoref{fig:sus-ecl-mrt}), supporting H2a and H2b.
MRT did not significantly predict ICL or GCL.

A complementary power analysis indicated the two significant MRT effects were detected with high power (ECL: $1-\beta=.91$; SUS: $1-\beta=.92$ at $\alpha=.05$; $.77$ and $.80$ under the Bonferroni-adjusted threshold), whereas power for the non-significant effects was low (ICL: $1-\beta=.48$; GCL: $1-\beta=.28$). 
These null findings should be interpreted with caution, as small associations cannot be ruled out.

\subsection{Spatial ability across instructional conditions}
\label{results-RQ3}

\begin{figure}[ht]
    \centering
    \begin{subfigure}[t]{0.45\linewidth}
        \centering
        \includegraphics[width=\linewidth]{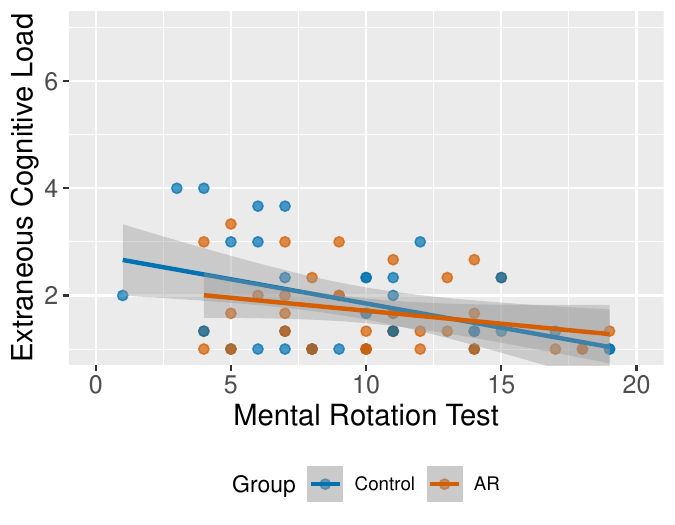}
        \caption{}
        \label{fig:rq3-ecl-moderation}
    \end{subfigure}\hfill
    \begin{subfigure}[t]{0.45\linewidth}
        \centering
        \includegraphics[width=\linewidth]{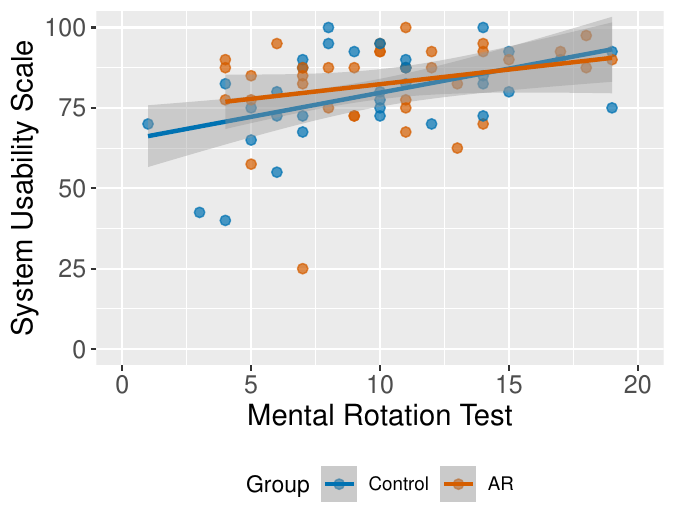}
        \caption{}
        \label{fig:rq3-sus-moderation}
    \end{subfigure}
    \caption{RQ3 moderation: ECL (a) and system usability (b) by MRT score, with separate regression lines and 95\% CIs for the AR and control conditions.}
    \Description{Two scatterplots comparing the AR and control conditions.
    Extraneous cognitive load decreases more strongly with Mental Rotation Test score in the control condition, while usability increases with score in both conditions and more strongly in the control condition.}
    \label{fig:rq3-moderation}
\end{figure}

Additional exploratory analyses examined whether AR moderates the influence of spatial ability on user experience via separate regression models for the AR and control conditions (\autoref{tab:rq3-moderation}; \autoref{fig:rq3-moderation}).
In the control condition, MRT significantly predicted ECL and SUS (both $p<.05$), with higher spatial ability linked to lower ECL and higher usability, mirroring the full sample, but did not significantly predict ICL or GCL.
In the AR condition, none of these associations remained significant after Bonferroni correction.
For GCL, the unadjusted association was significant, $F(1,34)=5.31$, $p=.027$, $R^2=.14$, but did not remain significant after correction ($p_{\mathrm{adj}}=.110$).

\begin{table}[ht]
    \centering
    \caption{Exploratory moderation summary (RQ3): MRT score as predictor within each experimental condition.}
    \label{tab:rq3-moderation}
    \resizebox{0.66\columnwidth}{!}{%
    \begin{tabular}{l
                    S[table-format=-1.2]
                    S[table-format=-1.2]
                    S[table-format=1.3]
                    S[table-format=1.2]
                    S[table-format=-1.2]
                    S[table-format=-1.2]
                    S[table-format=1.3]
                    S[table-format=1.2]}
        \toprule
        & \multicolumn{4}{c}{\textbf{Control} ($n=35$)} & \multicolumn{4}{c}{\textbf{AR} ($n=36$)} \\
        \cmidrule(lr){2-5} \cmidrule(lr){6-9}
        \textbf{Outcome} & {$b$} & {$\beta$} & {$p$} & {$\eta^2_p$} & {$b$} & {$\beta$} & {$p$} & {$\eta^2_p$} \\
        \midrule
        ECL & -0.09 & -0.42 & .047 & 0.18 & -0.05 & -0.29 & .424 & 0.08 \\
        ICL & -0.08 & -0.26 & .532 & 0.07 & -0.03 & -0.14 & 1.000 & 0.02 \\
        GCL & 0.00 & 0.00 & 1.000 & 0.00 & 0.07 & 0.37 & .110 & 0.14 \\
        SUS & 1.50 & 0.48 & .016 & 0.23 & 0.91 & 0.27 & .452 & 0.07 \\
        \bottomrule
    \end{tabular}
    }
    \vspace{1ex}

    \raggedright
    \footnotesize
    Note.
    Separate regressions were estimated within each condition, with MRT as predictor; $p$-values were Bonferroni-adjusted within condition ($\times4$).
    ECL/ICL/GCL = extraneous/intrinsic/germane cognitive load; SUS = System Usability Scale.
\end{table}

\section{Discussion}
\label{Discussion}

This study examined how AR and spatial ability relate to user experience during robot programming.
Higher spatial ability was associated with lower ECL and higher usability, whereas the AR condition provided no measurable group-level advantage over the control condition.
Exploratory analyses indicated weaker associations with spatial ability in the AR condition.
For HRI, these findings show why evaluations of robot-programming interfaces should consider user characteristics alongside average interface effects.

\subsection{Spatial ability and user experience in robot programming}
\label{spatial-ability-effect}

RQ2 examined how spatial ability relates to user experience during robot programming.
Both H2a and H2b were supported: higher spatial ability was associated with lower ECL and higher system usability.
Together, the ECL and SUS findings suggest that spatial ability relates to users' perceived processing demands and their evaluation of the complete robot-programming system.
When users must mentally translate coordinate systems, joint rotations, and interface commands into physical robot movements, difficulty with these translations may be experienced as extraneous load and lower system usability.
Previous studies have examined spatial ability mainly in AR-supported anatomy learning and other comparatively static visualization tasks \cite{kruger_learning_2022,ho_role_2022,bogomolova_effect_2020}.
The present findings expand evidence for the relevance of spatial ability to dynamic interaction with a physical robot.
This responds to calls to consider user characteristics more explicitly in AR research \cite{kozlova_bringing_2025,gonnermann-muller_unlocking_2025}.

ICL and GCL were included to characterize cognitive load, and we did not formulate directional hypotheses for either dimension.
Because intrinsic load depends on the demands a task places on users' existing capabilities, higher spatial ability could have reduced ICL in this spatially demanding task.
MRT was not significantly associated with ICL, although small effects cannot be ruled out.
The absence of an MRT association with ICL suggests that spatial ability was related more to demands arising from the interface and instructional format than to the perceived intrinsic complexity of the task.
MRT was also not significantly associated with GCL, but the low internal consistency of the GCL subscale limits interpretation of this null result.

Overall, the role of spatial ability was most evident in lower ECL and higher system usability, suggesting a specific relationship with interface-related demands and system evaluation rather than with cognitive load generally.

\subsection{Spatial ability across instructional conditions}
\label{moderation-effect}

RQ3 explored whether AR moderates the relationship between spatial ability and user experience via exploratory subgroup analyses comparing MRT associations with ECL, ICL, GCL, and SUS across conditions.
In the control condition, higher spatial ability was linked to lower ECL and higher SUS, even more so than in the overall sample.
These associations disappeared in the AR condition.
ICL and GCL showed similar trends across both conditions.

This pattern is consistent with the ability-as-compensator account and with studies reporting that AR reduced differences associated with spatial ability \cite{bogomolova_effect_2020,ho_effects_2024}.
It differs from studies supporting the ability-as-enhancer account \cite{kruger_learning_2022,ho_role_2022} and from work finding no role of spatial ability \cite{weng_effect_2023}.
These differences across studies may reflect the spatial demands of the respective tasks and the form of visualization provided.

This suggests that AR provided a comparably equitable user experience regardless of spatial ability, whereas spatial ability remained a significant predictor in the control condition.
We interpret this pattern as consistent with AR compensating for lower spatial ability.

This compensatory pattern is consistent with the AR application's design, which drew on \textit{Spatiality} and \textit{Contextuality} to present coordinate systems, joint rotations, and waypoints at their corresponding physical locations on or around the robot.
These visualizations may have reduced the need to integrate information mentally across instructions, controls, and robot movement, consistent with the split-attention principle \cite{ayres_split-attention_2014,mayer_multimedia_2020}.
By externalizing spatial relations that users would otherwise have to construct mentally, the interface may have reduced the demands placed on users with lower spatial ability.
The exploratory analyses cannot determine whether this pattern primarily reflects support for users with lower spatial ability or less additional value for users with higher spatial ability.

Instructors and system designers should consider users' spatial ability when deciding whether to deploy AR-based spatial scaffolding, rather than adopting AR as a default enhancement for everyone.
This aligns with Industry 5.0's emphasis on human-centric manufacturing and inclusive workforce development \cite{chang_survey_2024}.
Longitudinal designs are also needed to determine whether such compensatory effects persist as users gain expertise or diminish once scaffolding is no longer required.

\subsection{No overall advantage of AR over conventional instruction}
\label{improving-learningexp}

RQ1 examined whether the AR application improved user experience compared with the conventional system.
We hypothesized that the AR condition would reduce ECL (H1a) and increase system usability (H1b).
Neither H1a nor H1b was supported.
The instructional condition also did not significantly predict ICL or GCL.
The AR application therefore provided no measurable group-level advantage over the conventional approach.

This result contrasts with broader AR research reporting benefits for cognitive processes \cite{cai_effects_2021, lin_meta-analysis_2023}, as well as HRI findings on improved usability \cite{tsamis_intuitive_2021, yang_har2bot_2024} and reduced cognitive load \cite{yang_har2bot_2024}.
Rather than indicating a fundamental limitation of AR, the result provides information about the conditions under which AR may offer user-experience benefits.

Two features of the comparison help contextualize this result.
First, the AR application isolated spatial augmentation while keeping instructional content and controls comparable across conditions.
The finding therefore suggests that spatial augmentation alone, as implemented in this study, may not add measurable user-experience benefits over a well-designed conventional system \cite{kruger_augmented_2019}.
Second, the control condition already provided high usability, leaving limited room for further improvement.
AR systems that combine spatial augmentation with dynamic guidance, feedback, or additional interaction techniques may produce different outcomes \cite{ikeda_programar_2024,ong_augmented_2020,tsamis_intuitive_2021,yang_har2bot_2024,walker_teleoperation_2019}.

These results also inform the broader discussion on media comparison studies, which have been criticized for overemphasizing technology while underestimating instructional design \cite{buchner_media_2023}.
AR was not uniformly superior to a well-designed conventional method, and compensatory benefits instead appeared contingent on user characteristics.
Future research should move beyond simple media comparisons and examine how AR design features interact with user profiles, novelty, and baseline interface quality.
Controlled experiments that isolate specific features, such as interactive guidance, real-time feedback, or novel input techniques, can clarify which elements most effectively improve user experience.

\subsection{Limitations}
\label{Limitations}

Several methodological limitations must be acknowledged.

First, our exploratory analysis of AR's moderating effect (RQ3) relied on separate subgroup regressions rather than a single model with a condition $\times$ MRT interaction term, as the sample size did not provide sufficient power to reliably detect this interaction directly.

Second, we assessed cognitive load solely via self-report. 
Although the scale by \citet{klepsch_development_2017} is validated and widely used, self-assessments remain susceptible to subjective bias, and distinguishing between load types (ICL, ECL, GCL) is inherently challenging. 
Here, the GCL subscale showed relatively low internal consistency. 
Future studies could triangulate self-report data with objective measures like eye tracking or EEG.

Third, external validity is constrained by the study context: we evaluated video-see-through AR via a head-mounted display in a spatially demanding robot programming task, and these patterns may not generalize to handheld or screen-based AR, nor to tasks with lower spatial requirements, where spatial ability and augmentation may matter less.

\section{Conclusion}
\label{Conclusion}

This study examined how AR and spatial ability relate to user experience during programming a physical robot.
Higher spatial ability was associated with lower extraneous cognitive load and higher system usability, whereas the AR system showed no group-level advantage over conventional instruction.
Exploratory condition-specific analyses indicated a compensatory pattern: spatial ability predicted extraneous cognitive load and usability in the conventional condition but not in the AR condition.
AR may therefore support users with lower spatial ability by externalizing spatial relationships, even though it provided no uniform average benefit.
These findings motivate HRI research on targeted or adaptive spatial support and on how specific AR features interact with user characteristics.

\bibliographystyle{plainnat}
\bibliography{references}

@article{ajaykumar_survey_2021,
	author = {Ajaykumar, Gopika and Steele, Maureen and Huang, Chien-Ming},
	title = {A Survey on End-User Robot Programming},
	year = {2021},
	issue_date = {November 2022},
	publisher = {Association for Computing Machinery},
	address = {New York, NY, USA},
	volume = {54},
	number = {8},
	issn = {0360-0300},
	url = {https://doi.org/10.1145/3466819},
	doi = {10.1145/3466819},
	journal = {ACM Comput. Surv.},
	month = oct,
	articleno = {164},
	numpages = {36}
}

@article{ardito_approach_2006,
	title = {An approach to usability evaluation of e-learning applications},
	volume = {4},
	issn = {1615-5297},
	url = {https://doi.org/10.1007/s10209-005-0008-6},
	doi = {10.1007/s10209-005-0008-6},
	language = {en},
	number = {3},
	urldate = {2025-06-25},
	journal = {Universal Access in the Information Society},
	author = {Ardito, C. and Costabile, M. F. and Marsico, M. De and Lanzilotti, R. and Levialdi, S. and Roselli, T. and Rossano, V.},
	month = mar,
	year = {2006},
	pages = {270--283},
}

@incollection{ayres_split-attention_2014,
	edition = {2},
	series = {Cambridge handbooks in psychology},
	title = {The split-attention principle in multimedia learning},
	isbn = {978-1-107-61031-6 978-1-107-03520-1 978-1-139-99016-5},
	url = {https://doi.org/10.1017/CBO9781139547369.011},
	booktitle = {The {Cambridge} handbook of multimedia learning},
	publisher = {Cambridge University Press},
	author = {Ayres, Paul and Sweller, John},
	editor = {Mayer, Richard E},
	year = {2014},
	pages = {206--226},
}

@article{azuma_survey_1997,
	title = {A {Survey} of {Augmented} {Reality}},
	volume = {6},
	number = {4},
	doi = {10.1162/pres.1997.6.4.355},
	language = {en},
	journal = {Presence: Teleoperators and Virtual Environments},
	author = {Azuma, Ronald T},
	year = {1997},
	pages = {355--385},
}

@article{bogomolova_effect_2020,
	title = {The {Effect} of {Stereoscopic} {Augmented} {Reality} {Visualization} on {Learning} {Anatomy} and the {Modifying} {Effect} of {Visual}-{Spatial} {Abilities}: {A} {Double}-{Center} {Randomized} {Controlled} {Trial}},
	volume = {13},
	copyright = {© 2019 The Authors. Anatomical Sciences Education published by Wiley Periodicals, Inc. on behalf of American Association of Anatomists},
	issn = {1935-9780},
	shorttitle = {The {Effect} of {Stereoscopic} {Augmented} {Reality} {Visualization} on {Learning} {Anatomy} and the {Modifying} {Effect} of {Visual}-{Spatial} {Abilities}},
	url = {https://onlinelibrary.wiley.com/doi/abs/10.1002/ase.1941},
	doi = {10.1002/ase.1941},
	language = {en},
	number = {5},
	urldate = {2025-08-07},
	journal = {Anatomical Sciences Education},
	author = {Bogomolova, Katerina and van der Ham, Ineke J.M. and Dankbaar, Mary E.W. and van den Broek, Walter W. and Hovius, Steven E.R. and van der Hage, Jos A. and Hierck, Beerend P.},
	year = {2020},
	pages = {558--567},
}

@incollection{brooke_sus_1996,
	address = {London},
	title = {{SUS}: {A} '{Quick} and {Dirty}' {Usability} {Scale}},
	isbn = {978-0-429-15701-1},
	url = {https://doi.org/10.1201/9781498710411-35},
	booktitle = {Usability {Evaluation} {In} {Industry}},
	publisher = {CRC Press},
	author = {Brooke, John},
	editor = {Jordan, P. W. and Thomas, B. and Weerdmeester, B. A. and McClelland, I. L.},
	year = {1996},
	pages = {189--194},
}

@article{buchner_media_2023,
	title = {Media comparison studies dominate comparative research on augmented reality in education},
	volume = {195},
	issn = {0360-1315},
	url = {https://www.sciencedirect.com/science/article/pii/S0360131522002822},
	doi = {10.1016/j.compedu.2022.104711},
	urldate = {2025-07-07},
	journal = {Computers \& Education},
	author = {Buchner, Josef and Kerres, Michael},
	month = apr,
	year = {2023},
	pages = {104711},
}

@article{cai_effects_2021,
	title = {Effects of learning physics using {Augmented} {Reality} on students’ self-efficacy and conceptions of learning},
	volume = {52},
	issn = {1467-8535},
	url = {https://onlinelibrary.wiley.com/doi/abs/10.1111/bjet.13020},
	doi = {10.1111/bjet.13020},
	language = {en},
	number = {1},
	urldate = {2024-11-01},
	journal = {British Journal of Educational Technology},
	author = {Cai, Su and Liu, Changhao and Wang, Tao and Liu, Enrui and Liang, Jyh-Chong},
	year = {2021},
	pages = {235--251},
}

@article{carmigniani_augmented_2011,
	title = {Augmented reality technologies, systems and applications},
	volume = {51},
	issn = {1573-7721},
	url = {https://doi.org/10.1007/s11042-010-0660-6},
	doi = {10.1007/s11042-010-0660-6},
	number = {1},
	journal = {Multimedia Tools and Applications},
	author = {Carmigniani, Julie and Furht, Borko and Anisetti, Marco and Ceravolo, Paolo and Damiani, Ernesto and Ivkovic, Misa},
	month = jan,
	year = {2011},
	pages = {341--377},
}

@book{carroll_human_1993,
	address = {Cambridge},
	series = {Human cognitive abilities:  {A} survey of factor-analytic studies},
	title = {Human cognitive abilities:  {A} survey of factor-analytic studies},
	isbn = {978-0-521-38275-5 978-0-521-38712-5},
	shorttitle = {Human cognitive abilities},
	url = {https://doi.org/10.1017/CBO9780511571312},
	publisher = {Cambridge University Press},
	author = {Carroll, John B.},
	year = {1993},
}

@incollection{castro-alonso_sex_2019,
	address = {Cham},
	title = {Sex {Differences} in {Visuospatial} {Processing}},
	isbn = {978-3-030-20969-8},
	url = {https://doi.org/10.1007/978-3-030-20969-8_4},
	doi = {10.1007/978-3-030-20969-8_4},
	language = {en},
	urldate = {2025-06-10},
	booktitle = {Visuospatial {Processing} for {Education} in {Health} and {Natural} {Sciences}},
	publisher = {Springer},
	author = {Castro-Alonso, Juan C. and Jansen, Petra},
	editor = {Castro-Alonso, Juan C.},
	year = {2019},
	pages = {81--110},
}

@article{chandler_cognitive_1996,
	title = {Cognitive {Load} {While} {Learning} to {Use} a {Computer} {Program}},
	volume = {10},
	copyright = {Copyright © 1996 John Wiley \& Sons, Ltd.},
	issn = {1099-0720},
	url = {https://onlinelibrary.wiley.com/doi/abs/10.1002/%28SICI%291099-0720%28199604%2910%3A2%3C151%3A%3AAID-ACP380%3E3.0.CO%3B2-U},
	doi = {10.1002/(SICI)1099-0720(199604)10:2<151::AID-ACP380>3.0.CO;2-U},
	language = {en},
	number = {2},
	urldate = {2025-06-24},
	journal = {Applied Cognitive Psychology},
	author = {Chandler, Paul and Sweller, John},
	year = {1996},
	pages = {151--170},
}

@article{chang_survey_2024,
	title = {A {Survey} of {Augmented} {Reality} for {Human}–{Robot} {Collaboration}},
	volume = {12},
	copyright = {http://creativecommons.org/licenses/by/3.0/},
	issn = {2075-1702},
	url = {https://www.mdpi.com/2075-1702/12/8/540},
	doi = {10.3390/machines12080540},
	language = {en},
	number = {8},
	urldate = {2024-09-26},
	journal = {Machines},
	author = {Chang, Christine T. and Hayes, Bradley},
	month = aug,
	year = {2024},
	pages = {540},
}

@article{ericsson_verbal_1980,
	title = {Verbal reports as data},
	volume = {87},
	issn = {1939-1471},
	doi = {10.1037/0033-295X.87.3.215},
	number = {3},
	journal = {Psychological Review},
	author = {Ericsson, K. Anders and Simon, Herbert A.},
	year = {1980},
	pages = {215--251},
}

@article{franke_personal_2018,
	title = {A {Personal} {Resource} for {Technology} {Interaction}: {Development} and {Validation} of the {Affinity} for {Technology} {Interaction} ({ATI}) {Scale}},
	volume = {35},
	number = {6},
	doi = {10.1080/10447318.2018.1456150},
	journal = {International Journal of Human–Computer Interaction},
	author = {Franke, Thomas and Attig, Christiane and Wessel, Daniel},
	year = {2018},
	pages = {456--467},
}

@article{gao_multi-language_2020,
	title = {Multi-{Language} {Toolkit} for the {System} {Usability} {Scale}},
	volume = {36},
	issn = {1044-7318},
	url = {https://doi.org/10.1080/10447318.2020.1801173},
	doi = {10.1080/10447318.2020.1801173},
	number = {20},
	urldate = {2025-02-08},
	journal = {International Journal of Human–Computer Interaction},
	author = {Gao, Meiyuzi and Kortum, Philip and Oswald, Frederick L.},
	month = dec,
	year = {2020},
	pages = {1883--1901},
}

@article{gonnermann-muller_unlocking_2025,
	title = {Unlocking {Augmented} {Reality} {Learning} {Design} {Based} on {Evidence} {From} {Empirical} {Cognitive} {Load} {Studies} - {A} {Systematic} {Literature} {Review}},
	volume = {41},
	issn = {1365-2729},
	url = {https://onlinelibrary.wiley.com/doi/abs/10.1111/jcal.13095},
	doi = {10.1111/jcal.13095},
	language = {en},
	number = {1},
	urldate = {2024-12-04},
	journal = {Journal of Computer Assisted Learning},
	author = {Gonnermann-Müller, Jana and Krüger, Jule M.},
	year = {2025},
	pages = {e13095},
}

@inproceedings{hedayati_improving_2018,
	author = {Hedayati, Hooman and Walker, Michael and Szafir, Daniel},
	title = {Improving Collocated Robot Teleoperation with Augmented Reality},
	year = {2018},
	isbn = {9781450349536},
	publisher = {Association for Computing Machinery},
	address = {New York, NY, USA},
	url = {https://doi.org/10.1145/3171221.3171251},
	doi = {10.1145/3171221.3171251},
	booktitle = {Proceedings of the 2018 ACM/IEEE International Conference on Human-Robot Interaction},
	pages = {78--86},
	numpages = {9},
	location = {Chicago, IL, USA},
	series = {HRI '18}
}

@article{ho_effects_2024,
	title = {Effects of training methods on performance of a scaffolding task for workers with different spatial ability},
	volume = {30},
	issn = {1080-3548},
	url = {https://doi.org/10.1080/10803548.2024.2370643},
	doi = {10.1080/10803548.2024.2370643},
	number = {3},
	urldate = {2024-11-22},
	journal = {International Journal of Occupational Safety and Ergonomics},
	author = {Ho, Hung Yu and Wang, An Hsiang and Wu, Chia Huang},
	month = jul,
	year = {2024},
	pages = {985--994},
}

@article{ho_role_2022,
	title = {The role of spatial ability in mixed reality learning with the {HoloLens}},
	volume = {15},
	copyright = {© 2021 American Association for Anatomy},
	issn = {1935-9780},
	url = {https://onlinelibrary.wiley.com/doi/abs/10.1002/ase.2146},
	doi = {10.1002/ase.2146},
	language = {en},
	number = {6},
	urldate = {2024-11-22},
	journal = {Anatomical Sciences Education},
	author = {Ho, Simon and Liu, Pu and Palombo, Daniela J. and Handy, Todd C. and Krebs, Claudia},
	year = {2022},
	pages = {1074--1085},
}

@article{holmes_move_2018,
	title = {Move to learn: {Integrating} spatial information from multiple viewpoints},
	volume = {178},
	issn = {0010-0277},
	shorttitle = {Move to learn},
	url = {https://www.sciencedirect.com/science/article/pii/S0010027718301215},
	doi = {10.1016/j.cognition.2018.05.003},
	urldate = {2025-06-17},
	journal = {Cognition},
	author = {Holmes, Corinne A. and Newcombe, Nora S. and Shipley, Thomas F.},
	month = sep,
	year = {2018},
	pages = {7--25},
}

@article{howard_meta-analysis_2023,
	title = {A {Meta}-analysis of augmented reality programs for education and training},
	issn = {1434-9957},
	url = {https://doi.org/10.1007/s10055-023-00844-6},
	doi = {10.1007/s10055-023-00844-6},
	journal = {Virtual Reality},
	author = {Howard, Matt C. and Davis, Maggie M.},
	month = aug,
	year = {2023},
}

@article{huk_who_2006,
	title = {Who benefits from learning with {3D} models? the case of spatial ability},
	volume = {22},
	issn = {1365-2729},
	shorttitle = {Who benefits from learning with {3D} models?},
	url = {https://onlinelibrary.wiley.com/doi/abs/10.1111/j.1365-2729.2006.00180.x},
	doi = {10.1111/j.1365-2729.2006.00180.x},
	language = {en},
	number = {6},
	urldate = {2025-06-28},
	journal = {Journal of Computer Assisted Learning},
	author = {Huk, T.},
	year = {2006},
	pages = {392--404},
}

@inproceedings{ikeda_marcer_2025,
	author = {Ikeda, Bryce and Gramopadhye, Maitrey and Nekervis, LillyAnn and Szafir, Daniel},
	title = {{MARCER}: {Multimodal} {Augmented} {Reality} for {Composing} and {Executing} {Robot} {Tasks}},
	year = {2025},
	publisher = {IEEE Press},
	booktitle = {Proceedings of the 2025 {ACM}/{IEEE} {International} {Conference} on {Human}-{Robot} {Interaction}},
	pages = {529--539},
	numpages = {11},
	location = {Melbourne, Australia},
	series = {{HRI} '25},
}

@article{ikeda_programar_2024,
	title = {{PRogramAR}: {Augmented} {Reality} {End}-{User} {Robot} {Programming}},
	volume = {13},
	shorttitle = {{PRogramAR}},
	url = {https://dl.acm.org/doi/10.1145/3640008},
	doi = {10.1145/3640008},
	number = {1},
	urldate = {2024-09-26},
	journal = {J. Hum.-Robot Interact.},
	author = {Ikeda, Bryce and Szafir, Daniel},
	year = {2024},
	pages = {1--20},
}

@article{ivanov_new_2018,
	title = {New flexibility drivers for manufacturing, supply chain and service operations},
	volume = {56},
	number = {10},
	pages = {3359--3368},
	issn = {0020-7543},
	url = {https://doi.org/10.1080/00207543.2018.1457813},
	doi = {10.1080/00207543.2018.1457813},
	language = {EN},
	urldate = {2024-11-29},
	journal = {International Journal of Production Research},
	publisher = {Taylor \& Francis},
	author = {Ivanov, Dmitry and Das, Ajay and Choi, Tsan-Ming},
	month = may,
	year = {2018},
}

@article{jerabek_perceptual_2015,
	series = {The {Proceedings} of 6th {World} {Conference} on educational {Sciences}},
	title = {Perceptual {Specifics} and {Categorisation} of {Augmented} {Reality} {Systems}},
	volume = {191},
	issn = {1877-0428},
	url = {https://www.sciencedirect.com/science/article/pii/S1877042815026798},
	doi = {10.1016/j.sbspro.2015.04.419},
	urldate = {2025-06-17},
	journal = {Procedia - Social and Behavioral Sciences},
	author = {Jeřábek, Tomáš and Rambousek, Vladimír and Wildová, Radka},
	month = jun,
	year = {2015},
	pages = {1740--1744},
}

@inproceedings{jiang_comprehensive_2024,
	author = {Jiang, Xinkai and Mattes, Paul and Jia, Xiaogang and Schreiber, Nicolas and Neumann, Gerhard and Lioutikov, Rudolf},
	title = {A Comprehensive User Study on Augmented Reality-Based Data Collection Interfaces for Robot Learning},
	year = {2024},
	isbn = {9798400703225},
	publisher = {Association for Computing Machinery},
	address = {New York, NY, USA},
	url = {https://doi.org/10.1145/3610977.3634995},
	doi = {10.1145/3610977.3634995},
	booktitle = {Proceedings of the 2024 ACM/IEEE International Conference on Human-Robot Interaction},
	pages = {333--342},
	numpages = {10},
	location = {Boulder, CO, USA},
	series = {HRI '24}
}

@article{johnson-glenberg_embodied_2017,
	title = {Embodied science and mixed reality: {How} gesture and motion capture affect physics education},
	volume = {2},
	issn = {2365-7464},
	shorttitle = {Embodied science and mixed reality},
	url = {https://doi.org/10.1186/s41235-017-0060-9},
	doi = {10.1186/s41235-017-0060-9},
	number = {1},
	urldate = {2025-06-17},
	journal = {Cognitive Research: Principles and Implications},
	author = {Johnson-Glenberg, Mina C. and Megowan-Romanowicz, Colleen},
	month = may,
	year = {2017},
	pages = {24},
}

@article{klepsch_development_2017,
	title = {Development and {Validation} of {Two} {Instruments} {Measuring} {Intrinsic}, {Extraneous}, and {Germane} {Cognitive} {Load}},
	volume = {8},
	issn = {1664-1078},
	url = {https://www.frontiersin.org/journals/psychology/articles/10.3389/fpsyg.2017.01997/full},
	doi = {10.3389/fpsyg.2017.01997},
	language = {English},
	urldate = {2024-11-06},
	journal = {Frontiers in Psychology},
	publisher = {Frontiers},
	author = {Klepsch, Melina and Schmitz, Florian and Seufert, Tina},
	month = nov,
	year = {2017},
}

@article{kozlova_bringing_2025,
	title = {Bringing learners into focus: {A} systematic review of learner characteristics in {AR}-supported {STEM} education},
	volume = {122},
	issn = {1041-6080},
	shorttitle = {Bringing learners into focus},
	url = {https://www.sciencedirect.com/science/article/pii/S1041608025001037},
	doi = {10.1016/j.lindif.2025.102727},
	urldate = {2025-07-31},
	journal = {Learning and Individual Differences},
	author = {Kozlova, Zoya and Bach, Katharina M. and Edelsbrunner, Peter A. and Hofer, Sarah I.},
	month = aug,
	year = {2025},
	pages = {102727},
}

@inproceedings{kruger_augmented_2019,
	title = {Augmented reality in education: three unique characteristics from a user’s perspective},
	doi = {10.58459/icce.2019.481},
	booktitle = {International {Conference} on {Computers} in {Education}},
	publisher = {Asia-Pacific Society for Computers in Education},
	author = {Krüger, Jule M. and Buchholz, Alexander and Bodemer, Daniel},
	year = {2019},
	pages = {412--422},
}

@article{kruger_learning_2022,
	title = {Learning with augmented reality: {Impact} of dimensionality and spatial abilities},
	volume = {3},
	issn = {26665573},
	shorttitle = {Learning with augmented reality},
	url = {https://linkinghub.elsevier.com/retrieve/pii/S2666557321000367},
	doi = {10.1016/j.caeo.2021.100065},
	language = {en},
	urldate = {2023-04-25},
	journal = {Computers and Education Open},
	author = {Krüger, Jule M. and Palzer, Kevin and Bodemer, Daniel},
	month = dec,
	year = {2022},
	pages = {100065},
}

@article{lee_augmented_2012,
	title = {Augmented {Reality} in {Education} and {Training}},
	volume = {56},
	issn = {1559-7075},
	url = {https://doi.org/10.1007/s11528-012-0559-3},
	doi = {10.1007/s11528-012-0559-3},
	language = {en},
	number = {2},
	urldate = {2025-06-22},
	journal = {TechTrends},
	author = {Lee, Kangdon},
	month = mar,
	year = {2012},
	pages = {13--21},
}

@article{lin_meta-analysis_2023,
	title = {A meta-analysis of the effects of augmented reality technologies in interactive learning environments (2012–2022)},
	volume = {31},
	issn = {1099-0542},
	url = {https://onlinelibrary.wiley.com/doi/abs/10.1002/cae.22628},
	doi = {10.1002/cae.22628},
	language = {en},
	number = {4},
	urldate = {2024-08-27},
	journal = {Computer Applications in Engineering Education},
	author = {Lin, Yupeng and Yu, Zhonggen},
	year = {2023},
	pages = {1111--1131},
}

@article{linn_emergence_1985,
	title = {Emergence and {Characterization} of {Sex} {Differences} in {Spatial} {Ability}: {A} {Meta}-{Analysis}},
	volume = {56},
	issn = {0009-3920},
	shorttitle = {Emergence and {Characterization} of {Sex} {Differences} in {Spatial} {Ability}},
	url = {https://www.jstor.org/stable/1130467},
	doi = {10.2307/1130467},
	number = {6},
	urldate = {2025-07-10},
	journal = {Child Development},
	author = {Linn, Marcia C. and Petersen, Anne C.},
	year = {1985},
	pages = {1479--1498},
}

@article{lotsaris_ar_2021,
	series = {8th {CIRP} {Conference} of {Assembly} {Technology} and {Systems}},
	title = {{AR} based robot programming using teaching by demonstration techniques},
	volume = {97},
	issn = {2212-8271},
	url = {https://www.sciencedirect.com/science/article/pii/S2212827120314906},
	doi = {10.1016/j.procir.2020.09.186},
	urldate = {2024-09-27},
	journal = {Procedia CIRP},
	author = {Lotsaris, Konstantinos and Gkournelos, Christos and Fousekis, Nikos and Kousi, Niki and Makris, Sotiris},
	month = jan,
	year = {2021},
	pages = {459--463},
}

@article{makris_augmented_2016,
	title = {Augmented reality system for operator support in human–robot collaborative assembly},
	volume = {65},
	issn = {0007-8506},
	url = {https://www.sciencedirect.com/science/article/pii/S0007850616300385},
	doi = {10.1016/j.cirp.2016.04.038},
	number = {1},
	urldate = {2024-09-28},
	journal = {CIRP Annals},
	author = {Makris, Sotiris and Karagiannis, Panagiotis and Koukas, Spyridon and Matthaiakis, Aleksandros-Stereos},
	month = jan,
	year = {2016},
	pages = {61--64},
}

@book{mayer_multimedia_2020,
	address = {Cambridge},
	edition = {3},
	title = {Multimedia {Learning}},
	doi = {10.1017/9781316941355},
	publisher = {Cambridge University Press},
	author = {Mayer, Richard E.},
	year = {2020},
}

@inproceedings{moorman_teaching_2026,
	author = {Moorman, Nina Marie and Luebbers, Matthew and Xi-Jia, Zhang and Lau, Yee Ching (Marcus) and Yao, Yixing and Langwasser, Megan and Zaidi, Zulfiqar and Chen, Letian and van Waveren, Sanne and Gombolay, Matthew},
	title = {Teaching the Teacher: Live Foundation Model and Augmented Reality Feedback for Human-to-Robot Skill Transfer},
	year = {2026},
	isbn = {9798400721281},
	publisher = {Association for Computing Machinery},
	address = {New York, NY, USA},
	url = {https://doi.org/10.1145/3757279.3788667},
	doi = {10.1145/3757279.3788667},
	booktitle = {Proceedings of the 21st ACM/IEEE International Conference on Human-Robot Interaction},
	pages = {544--552},
	numpages = {9},
	location = {Edinburgh, Scotland, UK},
	series = {HRI '26}
}

@incollection{moreno_cognitive_2010,
	address = {Cambridge},
	title = {Cognitive {Load} {Theory}: {Historical} {Development} and {Relation} to {Other} {Theories}},
	isbn = {978-0-521-86023-9},
	shorttitle = {Cognitive {Load} {Theory}},
	doi = {10.1017/CBO9780511844744.003},
	urldate = {2025-07-12},
	booktitle = {Cognitive {Load} {Theory}},
	publisher = {Cambridge University Press},
	author = {Moreno, Roxana and Park, Babette},
	editor = {Plass, Jan L. and Brünken, Roland and Moreno, Roxana},
	year = {2010},
	pages = {9--28},
}

@article{neves_application_2020,
	title = {Application of mixed reality in robot manipulator programming},
	volume = {45},
	issn = {0143-991X},
	url = {https://doi.org/10.1108/IR-06-2018-0120},
	doi = {10.1108/IR-06-2018-0120},
	number = {6},
	urldate = {2024-09-28},
	journal = {Industrial Robot: An International Journal},
	publisher = {Emerald Publishing Limited},
	author = {Neves, João and Serrario, Diogo and Pires, J. Norberto},
	year = {2018},
	pages = {784--793},
}

@misc{noauthor_polyscope_2026,
	title = {{PolyScope} 5 - {More} than just keep up},
	url = {https://www.universal-robots.com/products/polyscope-5/},
	language = {en},
	urldate = {2026-09-09},
	year = {2026},
}

@article{ong_augmented_2020,
	title = {Augmented reality-assisted robot programming system for industrial applications},
	volume = {61},
	issn = {0736-5845},
	url = {https://www.sciencedirect.com/science/article/pii/S0736584519300250},
	doi = {10.1016/j.rcim.2019.101820},
	urldate = {2024-09-20},
	journal = {Robotics and Computer-Integrated Manufacturing},
	author = {Ong, S. K. and Yew, A. W. W. and Thanigaivel, N. K. and Nee, A. Y. C.},
	month = feb,
	year = {2020},
	pages = {101820},
}

@article{peters_redrawn_1995,
	title = {A {Redrawn} {Vandenberg} and {Kuse} {Mental} {Rotations} {Test} - {Different} {Versions} and {Factors} {That} {Affect} {Performance}},
	volume = {28},
	issn = {0278-2626},
	url = {https://www.sciencedirect.com/science/article/pii/S0278262685710329},
	doi = {10.1006/brcg.1995.1032},
	number = {1},
	urldate = {2024-10-08},
	journal = {Brain and Cognition},
	author = {Peters, M. and Laeng, B. and Latham, K. and Jackson, M. and Zaiyouna, R. and Richardson, C.},
	month = jun,
	year = {1995},
	pages = {39--58},
}

@article{radu_augmented_2014,
	title = {Augmented reality in education: a meta-review and cross-media analysis},
	volume = {18},
	issn = {1617-4909, 1617-4917},
	shorttitle = {Augmented reality in education},
	url = {http://link.springer.com/10.1007/s00779-013-0747-y},
	doi = {10.1007/s00779-013-0747-y},
	language = {en},
	number = {6},
	urldate = {2021-12-05},
	journal = {Personal and Ubiquitous Computing},
	author = {Radu, Iulian},
	month = aug,
	year = {2014},
	pages = {1533--1543},
}

@inproceedings{sautter_mixed_2021,
	title = {Mixed {Reality} {Supported} {Learning} for {Industrial} on-the-job {Training}},
	url = {https://ssrn.com/abstract=3864189},
	doi = {10.2139/ssrn.3864189},
	language = {en},
	urldate = {2023-10-05},
	booktitle = {Proceedings of the Conference on Learning Factories ({CLF}) 2021},
	author = {Sautter, Björn and Daling, Lea},
	month = jun,
	year = {2021},
}

@article{sweller_cognitive_1998,
	title = {Cognitive architecture and instructional design},
	volume = {10},
	doi = {10.1023/A:1022193728205},
	number = {3},
	journal = {Educational psychology review},
	publisher = {Springer},
	author = {Sweller, John and Van Merrienboer, Jeroen JG and Paas, Fred G. W. C.},
	year = {1998},
	pages = {251--296},
}

@article{sweller_cognitive_2019,
	title = {Cognitive {Architecture} and {Instructional} {Design}: 20 {Years} {Later}},
	volume = {31},
	issn = {1040-726X, 1573-336X},
	shorttitle = {Cognitive {Architecture} and {Instructional} {Design}},
	url = {http://link.springer.com/10.1007/s10648-019-09465-5},
	doi = {10.1007/s10648-019-09465-5},
	language = {en},
	number = {2},
	urldate = {2022-01-20},
	journal = {Educational Psychology Review},
	author = {Sweller, John and van Merriënboer, Jeroen J. G. and Paas, Fred},
	month = jun,
	year = {2019},
	pages = {261--292},
}

@article{sweller_element_2010,
	title = {Element {Interactivity} and {Intrinsic}, {Extraneous}, and {Germane} {Cognitive} {Load}},
	volume = {22},
	issn = {1040-726X, 1573-336X},
	url = {http://link.springer.com/10.1007/s10648-010-9128-5},
	doi = {10.1007/s10648-010-9128-5},
	language = {en},
	number = {2},
	urldate = {2022-01-20},
	journal = {Educational Psychology Review},
	author = {Sweller, John},
	month = jun,
	year = {2010},
	pages = {123--138},
}

@inproceedings{tsamis_intuitive_2021,
	title = {Intuitive and {Safe} {Interaction} in {Multi}-{User} {Human} {Robot} {Collaboration} {Environments} through {Augmented} {Reality} {Displays}},
	issn = {1944-9437},
	url = {https://ieeexplore.ieee.org/document/9515474},
	doi = {10.1109/RO-MAN50785.2021.9515474},
	urldate = {2025-06-27},
	booktitle = {2021 30th {IEEE} {International} {Conference} on {Robot} \& {Human} {Interactive} {Communication} ({RO}-{MAN})},
	author = {Tsamis, Georgios and Chantziaras, Georgios and Giakoumis, Dimitrios and Kostavelis, Ioannis and Kargakos, Andreas and Tsakiris, Athanasios and Tzovaras, Dimitrios},
	month = aug,
	year = {2021},
	pages = {520--526},
}

@misc{unity_technologies_ros_2025,
	title = {{ROS} {TCP} {Endpoint}},
	copyright = {Apache-2.0},
	url = {https://github.com/Unity-Technologies/ROS-TCP-Endpoint},
	urldate = {2025-07-03},
	publisher = {GitHub},
	author = {{Unity Technologies}},
	year = {2025},
}

@misc{universal_robots_as_universal_2025,
	title = {Universal {Robots} {ROS2} {Driver}},
	copyright = {BSD-3-Clause},
	url = {https://github.com/UniversalRobots/Universal_Robots_ROS2_Driver},
	urldate = {2025-07-03},
	publisher = {GitHub},
	author = {{Universal Robots A/S}},
	year = {2025},
}

@article{vlachogianni_perceived_2022,
	title = {Perceived usability evaluation of educational technology using the {System} {Usability} {Scale} ({SUS}): {A} systematic review},
	volume = {54},
	number = {3},
	pages = {392--409},
	issn = {1539-1523},
	shorttitle = {Perceived usability evaluation of educational technology using the {System} {Usability} {Scale} ({SUS})},
	doi = {10.1080/15391523.2020.1867938},
	language = {EN},
	urldate = {2024-11-30},
	journal = {Journal of Research on Technology in Education},
	publisher = {Routledge},
	author = {Vlachogianni, Prokopia and Tselios, Nikolaos},
	month = jul,
	year = {2022},
}

@inproceedings{walker_communicating_2018,
	author = {Walker, Michael and Hedayati, Hooman and Lee, Jennifer and Szafir, Daniel},
	title = {Communicating Robot Motion Intent with Augmented Reality},
	year = {2018},
	isbn = {9781450349536},
	publisher = {Association for Computing Machinery},
	address = {New York, NY, USA},
	url = {https://doi.org/10.1145/3171221.3171253},
	doi = {10.1145/3171221.3171253},
	booktitle = {Proceedings of the 2018 ACM/IEEE International Conference on Human-Robot Interaction},
	pages = {316--324},
	numpages = {9},
	location = {Chicago, IL, USA},
	series = {HRI '18},
}

@INPROCEEDINGS{walker_teleoperation_2019,
  author={Walker, Michael E. and Hedayati, Hooman and Szafir, Daniel},
  booktitle={2019 14th ACM/IEEE International Conference on Human-Robot Interaction (HRI)}, 
  title={Robot Teleoperation with Augmented Reality Virtual Surrogates}, 
  year={2019},
  volume={},
  number={},
  pages={202-210},
  doi={10.1109/HRI.2019.8673306}}

@article{webel_augmented_2013,
	series = {Models and {Technologies} for {Multi}-modal {Skill} {Training}},
	title = {An augmented reality training platform for assembly and maintenance skills},
	volume = {61},
	issn = {0921-8890},
	url = {https://www.sciencedirect.com/science/article/pii/S0921889012001674},
	doi = {10.1016/j.robot.2012.09.013},
	number = {4},
	urldate = {2025-06-22},
	journal = {Robotics and Autonomous Systems},
	author = {Webel, Sabine and Bockholt, Uli and Engelke, Timo and Gavish, Nirit and Olbrich, Manuel and Preusche, Carsten},
	month = apr,
	year = {2013},
	pages = {398--403},
}

@article{weng_effect_2023,
	title = {The effect of using theodolite {3D} {AR} in teaching measurement error on learning outcomes and satisfaction of civil engineering students with different spatial ability},
	volume = {31},
	issn = {1049-4820},
	url = {https://doi.org/10.1080/10494820.2021.1898989},
	doi = {10.1080/10494820.2021.1898989},
	number = {5},
	urldate = {2024-11-22},
	journal = {Interactive Learning Environments},
	author = {Weng, Cathy and Puspitasari, Dani and Tran, Khanh Nguyen Phuong and Feng, Pei Jie and Awuor, Nicholas O. and Matere, Isaac Manyonge},
	month = jul,
	year = {2023},
	pages = {2722--2736},
}

@article{yang_har2bot_2024,
	title = {{HAR2bot}: a human-centered augmented reality robot programming method with the awareness of cognitive load},
	volume = {35},
	doi = {10.1007/s10845-023-02096-2},
	number = {5},
	journal = {Journal of Intelligent Manufacturing},
	author = {Yang, Wenhao and Xiao, Qinqin and Zhang, Yunbo},
	year = {2024},
	pages = {1985 -- 2003},
}

\end{document}